\documentclass[letterpaper, 10 pt, conference]{ieeeconf}

\IEEEoverridecommandlockouts
\usepackage{graphicx}
\usepackage{amsmath}
\usepackage{amssymb}
\usepackage{url}

\title{\LARGE \bf
Markerless Multi-Modal Autonomous Robotic Inspection of Large Space Structures}

\author{
Juan de Dios Alfaro$^{1}$, Arturo Ríos$^{1}$, David Rodríguez-Martínez$^{1}$ and Carlos Pérez-del-Pulgar$^{1}$%
\thanks{$^{1}$Space Robotics Laboratory, University Institute for Research in Mechatronics and Cyber-Physical Systems, Universidad de Málaga, Málaga, Spain.}%
\thanks{This work was supported by the European Space Agency and Sirin Orbital Systems within the ECHO2 project: End-to-End Control for Handling Operations in Orbit.}
}

\begin{document}

\maketitle
\thispagestyle{empty}
\pagestyle{empty}


\begin{abstract}

Future orbital infrastructures, such as deployable antennas, solar farms, and large orbital platforms will require autonomous inspection systems able to operate with limited prior knowledge and without cooperative markers. Current on-orbit servicing approaches often rely on predefined trajectories, standard interfaces, fiducial markers or accurate target models, which limits scalability for large, heterogeneous or partially unknown structures. This paper presents a markerless autonomous robotic inspection pipeline in which 3D reconstruction is used as an inspection-support representation. The system integrates a Kinova Gen~2 manipulator with an end-effector-mounted multimodal sensor head composed of an RGB-D camera, a thermal camera and a 2D LiDAR. The pipeline estimates an approximate inspection volume, generates viewpoints, plans collision-free motions with MoveIt, and synchronously records RGB-D images, thermal data, and robot poses in ROS~2. Candidate reconstruction methods were evaluated to select a practical method for this pipeline, with Nerfacto used for geometric reconstruction and Thermal-Nerfacto used to demonstrate thermal-aware rendering for inspection. Validation in a Gazebo-based simulator and preliminary laboratory tests reveal that the proposed system can autonomously acquire spatially coherent inspection data and produce reconstructions suitable for visual and geometric assessment, representing a step towards inspection of large non-cooperative space structures.

\end{abstract}


\section{INTRODUCTION}

The next generation of orbital infrastructures will likely involve increasingly large, modular, and complex space structures. Examples include future space stations, large antennas, deployable solar arrays, orbital construction platforms, and even data centres located in orbit. These systems are expected to require frequent inspection, maintenance, repair, and possible reconfiguration during their operational lifetime. As their size and complexity increase, relying exclusively on human operators, preplanned trajectories, or dedicated cooperative interfaces becomes progressively less scalable.

Current on-orbit servicing missions are typically designed around highly controlled operational assumptions. In many cases, the target spacecraft is known in advance, its geometry is available, and servicing operations rely on predefined docking interfaces, grasping points, fiducial markers, or carefully planned sequences. This approach is suitable for specific servicing missions, but it limits the flexibility required for future orbital environments where structures may be large, heterogeneous, partially unknown, or not specifically designed for robotic inspection.

Autonomous robotic inspection therefore requires perception-driven systems capable of operating with limited prior knowledge of the inspected structure. Instead of depending on artificial markers or dedicated inspection interfaces, the robot should estimate the approximate volume of the target, plan suitable viewpoints, acquire data from multiple perspectives, and generate an inspection-ready representation of the scene. In this context, 3D reconstruction methods are especially relevant because they can provide spatially coherent models that support visual and geometric inspection, coverage assessment, anomaly localization, and future repair planning. However, their deployment in space robotics raises challenges beyond reconstruction quality alone. Inspection systems must operate without cooperative markers or complete target models, estimate an actionable inspection volume, generate feasible viewpoints under kinematic and collision constraints, and acquire spatially consistent multimodal data. Challenging visual conditions, such as low-texture or repetitive surfaces, strong reflections, and illumination changes, introduce additional difficulties for perception and reconstruction.

The main contribution of this paper is the integration of 3D reconstruction into a markerless autonomous robotic inspection pipeline, where reconstruction is treated as an enabling tool for inspection. The proposed system combines a 7DoF Kinova Gen~2 manipulator, a multimodal sensor head composed of an Intel RealSense D435i RGB-D camera, an Optris PI thermal camera, and an RPLIDAR S3 2D LiDAR, camera intrinsic and Eye-in-Hand calibration, LiDAR-based inspection-volume estimation, automatic viewpoint generation, MoveIt-based motion planning, and synchronized RGB-D, thermal, and pose logging in ROS~2. A Gazebo-based simulator is used to test inspection strategies and detect planning limitations before execution, while physical laboratory experiments validate the sensor integration, calibration, acquisition, and reconstruction workflow. In addition, several state-of-the-art reconstruction methods were evaluated to select a suitable candidate for integration into the inspection bench.

\begin{figure*}[t]
    \centering
    \includegraphics[width=\textwidth]{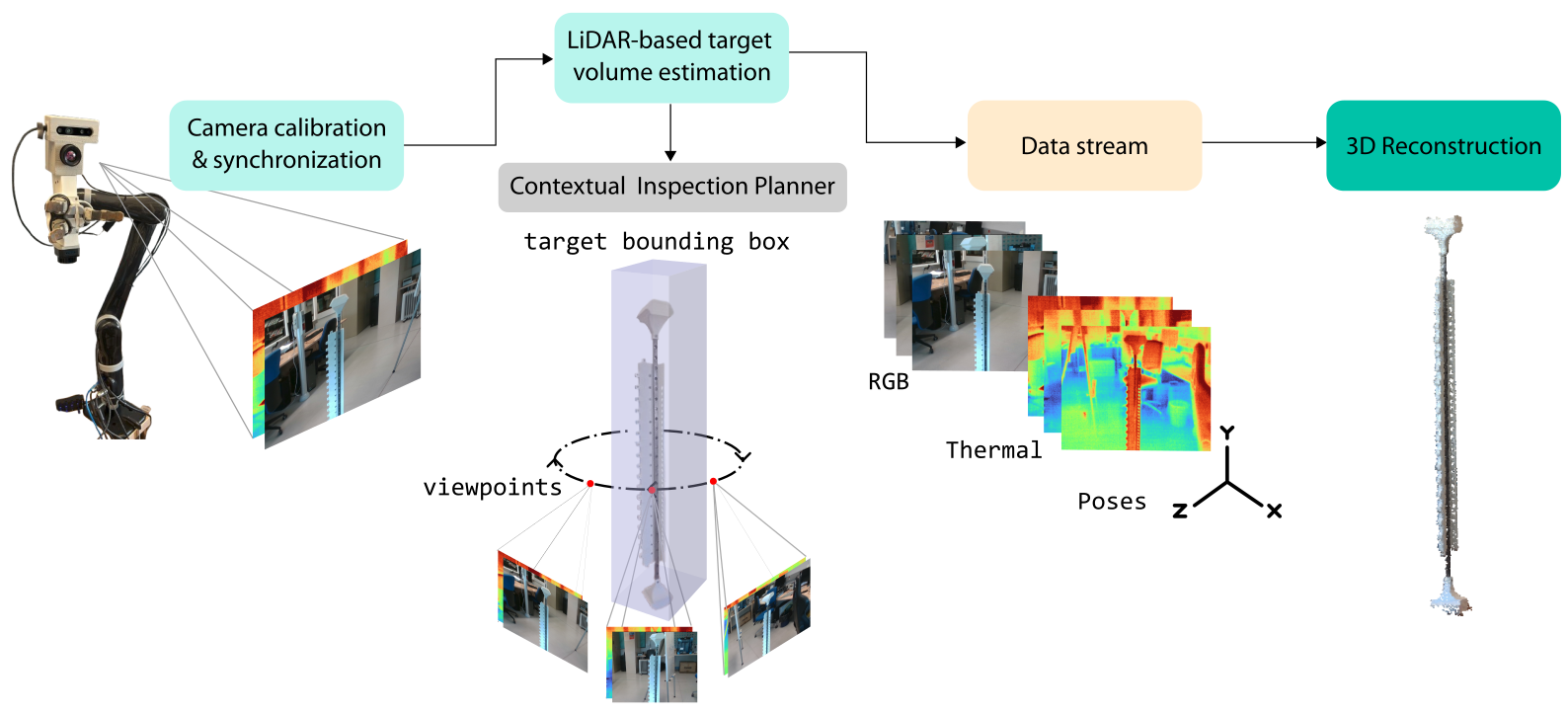}
    \caption{Overview of the proposed markerless autonomous robotic inspection pipeline. 
After camera calibration and synchronization, the system performs an initial LiDAR-based estimation of the target volume, generates inspection viewpoints around the resulting bounding box, plans collision-free motions, acquires synchronized RGB-D, thermal, and pose data, and finally reconstructs a 3D model for inspection purposes.}
    \label{fig:pipeline_overview}
\end{figure*}

\section{RELATED WORK}

\subsection{Robotic Inspection and On-Orbit Servicing}

Robotic inspection and on-orbit servicing are key technologies for extending the lifetime of satellites and enabling the assembly, maintenance, and recycling of future orbital infrastructures. Previous demonstrations, such as EROSS, have shown the feasibility of servicing operations in orbit~\cite{roa2024}. However, most of these approaches are still based on cooperative targets, predefined interfaces, known geometries, and carefully planned operational procedures.

This assumption is reasonable for current missions, where the target spacecraft is usually designed with servicing constraints in mind. However, future large space structures may involve modular elements, non-standard components, partial occlusions, variable illumination conditions, and complex geometries. In these situations, robotic systems must be able to adapt their inspection strategy using onboard perception and autonomous planning.

Several research initiatives have explored robotic assembly and manipulation in space, including modular spacecraft assembly, multi-arm robotic demonstrators, and orbital support services~\cite{letier2019}. Nevertheless, many of these approaches still depend on cooperative interfaces, previously defined grasping points, or external references. In contrast, the approach proposed in this work focuses on markerless inspection and autonomous data acquisition, using the geometry perceived by the robot to guide the inspection process.

\subsection{3D Reconstruction for Robotic Inspection}

Three-dimensional reconstruction plays a central role in autonomous inspection. A reconstructed model of the target structure can be used to analyse its geometry, verify coverage, inspect inaccessible areas through novel views, and provide a spatial basis for later diagnosis or repair planning.

Classical photogrammetry pipelines, such as Structure-from-Motion and Multi-View Stereo, have been widely used to recover sparse and dense 3D models from image collections. COLMAP is one of the most representative tools in this category, providing robust camera pose estimation and dense reconstruction when the scene contains sufficient texture and the image overlap is adequate~\cite{schonberger2016}. However, space-like environments can present challenging visual conditions, including low texture, strong reflections, harsh illumination changes, and repeated structural patterns.

More recently, neural rendering methods such as Neural Radiance Fields have shown strong potential for view synthesis and 3D scene representation~\cite{mildenhall2021}. NeRF-based methods learn a continuous volumetric representation of a scene from multiple posed images and can generate novel views with high visual quality. Nerfacto, implemented within the Nerfstudio framework, provides a practical and modular implementation suitable for robotic datasets~\cite{tancik2023}.

3D Gaussian Splatting enables high-quality real-time rendering, whereas neural implicit methods such as NeuS target smooth surface recovery when geometric precision is prioritised~\cite{kerbl2023,wang2021}.

CasMVSNet estimates depth from multi-view imagery, while MASt3R provides robust correspondences and 3D grounding~\cite{gu2020,leroy2024}. Their integration into an end-to-end robotic pipeline may require additional processing stages.

Recent works have also shown the potential of integrating 3D reconstruction methods into robotic workflows compatible with ROS~2~\cite{martinez2026}. In this work, these reconstruction families are not treated as isolated algorithms, but as candidates for integration into a robotic inspection system. Therefore, the selection criterion is not only reconstruction quality, but also robustness, ease of integration, computational feasibility, and compatibility with the generated RGB-T-D dataset.

\subsection{Multimodal Perception in Space Robotics}

Multimodal perception can increase the robustness of robotic inspection systems by combining complementary sources of information. RGB cameras provide rich visual appearance, depth sensors offer direct geometric measurements, LiDAR sensors can support spatial estimation and obstacle detection, and thermal cameras can reveal information that is not visible in the RGB spectrum.

Thermal perception is especially relevant for inspection tasks because many structural or electronic failures may appear as thermal anomalies before they are visible in the geometric or RGB domain. In orbital infrastructures, thermal gradients, overheating components, insulation defects, and material degradation can provide valuable diagnostic information. Previous work on thermal imagery for rover soil assessment under simulated Mars conditions shows the interest of incorporating thermal sensing into robotic systems for space-related environments~\cite{castilla2024}.

Although the geometric reconstruction experiments are mainly supported by RGB images and associated poses, this work also evaluates the use of synchronized thermal data through a thermal-aware neural reconstruction experiment. This makes the platform suitable not only for geometric modelling, but also for future work involving thermal mapping, anomaly detection, and combined geometric-thermal inspection.

\section{SYSTEM ARCHITECTURE}

\subsection{Overview}

The proposed system is organized as a complete markerless robotic inspection pipeline, as summarized in Fig.~\ref{fig:pipeline_overview}. The architecture combines perception, autonomous viewpoint generation, motion planning, synchronized multimodal data acquisition, and 3D reconstruction into a single ROS~2-based workflow.

The robotic platform consists of a Kinova Gen~2 manipulator equipped with a multimodal sensor head mounted on the end-effector, as shown in Fig.~\ref{fig:sensor_head}. This sensor head integrates an Intel RealSense D435i RGB-D camera, an Optris PI thermal camera, and an RPLIDAR S3 2D LiDAR. The LiDAR is used to estimate the approximate inspection volume of the target structure, while the RGB-D and thermal cameras acquire the visual, geometric, and thermal data required for inspection and reconstruction.

After camera calibration and Eye-in-Hand calibration, the system estimates a bounding box around the target structure, generates inspection viewpoints around this volume, orients the sensors towards the structure centroid, plans collision-free motions using MoveIt, and records synchronized RGB-D, thermal, and pose data. The resulting dataset is then used to reconstruct a 3D model of the inspected structure, which serves as the basis for visual and geometric inspection.

\subsection{Multimodal Sensor Head}

\begin{figure}[thpb]
    \centering
    \includegraphics[width=0.95\linewidth]{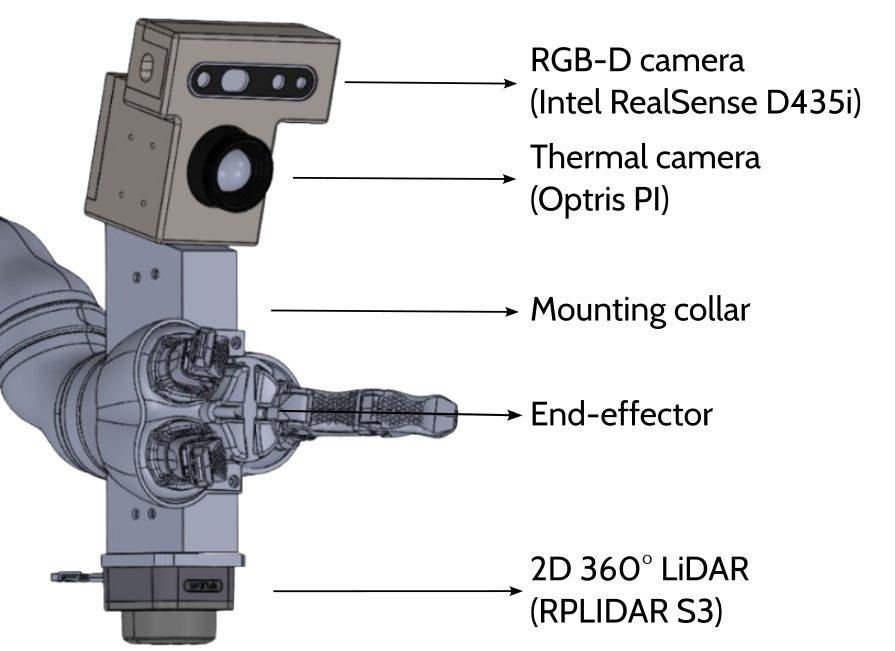}
    \caption{Main components of the multimodal sensor head mounted on the Kinova Gen~2 end-effector.}
    \label{fig:sensor_head}
\end{figure}

As shown in Fig.~\ref{fig:sensor_head}, the sensor head was designed to integrate different sensing modalities in a compact configuration. The sensor configuration was selected to provide complementary information throughout the inspection workflow. The RGB-D camera captures detailed visual appearance and depth information for reconstruction-oriented data acquisition. The 2D LiDAR is used during the initial exploration stage to obtain a coarse geometric estimate of the target volume, from which a practical bounding box can be derived for autonomous viewpoint generation; it is not intended as the primary reconstruction sensor. The thermal camera provides complementary radiometric information that can reveal temperature anomalies, overheating components, insulation defects, or material degradation that may not be visible in the RGB domain. In addition, thermal imagery could support perception in visually degraded regions where RGB information is less reliable due to low illumination, strong reflections, or insufficient texture.

A custom 3D-printed support was designed to attach the sensors to the end-effector of the Kinova Gen~2 manipulator. This design ensures a fixed relative configuration between sensors and allows the calibration parameters to remain valid during the acquisition process.

Other sensing modalities, such as radar and event cameras, are also relevant for space robotics but were not included in the present prototype. Radar is particularly suitable for illumination-independent range and velocity measurements at longer distances, whereas event-based cameras are advantageous under rapid relative motion and extreme illumination transients. In contrast, the proposed system targets close-range, stop-and-acquire inspection and image-based reconstruction. The selected RGB-D, LiDAR, and thermal configuration provides dense visual data for reconstruction, geometric information for viewpoint generation, and complementary thermal information for diagnosis, while maintaining a compact end-effector-mounted sensor head compatible with the implemented ROS 2 acquisition pipeline

\subsection{Gazebo-Based Simulation Environment}

A robotic setup was developed in Gazebo. The simulated environment includes the Kinova Gen~2 manipulator, the sensor head, and representative large space structure geometries. This virtual setup allows inspection strategies to be tested before execution on the physical robot.

The Gazebo environment enables repeatable validation of viewpoint generation and collision-free planning before execution on hardware, while revealing self-collision and workspace limitations.

In the simulated experiment, the manipulator is mounted on a holonomic mobile base. This does not aim to reproduce the full dynamics of a free-floating spacecraft, but rather to approximate the planar relative mobility available in terrestrial microgravity test benches. This configuration is coherent with future validation in orbital robotics laboratories where motion is restricted to a plane.

\section{CALIBRATION AND DATA ACQUISITION}

\subsection{Camera Calibration}

Camera intrinsic calibration is required to correctly interpret image measurements and to use the acquired images in 3D reconstruction pipelines. The intrinsic parameters of the RGB camera are estimated using a standard calibration procedure based on planar calibration patterns~\cite{zhang2000}. These parameters include focal lengths, principal point, and distortion coefficients.

Accurate intrinsics are essential for reconstruction consistency when robot-derived poses are used instead of poses estimated by the reconstruction software.

\subsection{Eye-in-Hand Calibration}

Since the camera is mounted on the robot end-effector, it is necessary to estimate the rigid transformation between the end-effector frame and the optical frame of the camera. This is known as Eye-in-Hand calibration.

The calibration problem is commonly formulated as:

\begin{equation}
    AX = XB
\end{equation}

where $A$ represents the relative motion of the robot end-effector between two poses, $B$ represents the relative motion observed by the camera with respect to a calibration marker, and $X$ is the unknown rigid transformation between the end-effector and the camera~\cite{tsai1989,park1994}.

In this work, an ArUco marker is used only during the calibration stage~\cite{garrido2014}. The robot moves through several known poses while the camera observes the marker. From these observations, the rigid transformation between the end-effector and the camera is estimated using an Eye-in-Hand calibration procedure based on easy\_handeye~\cite{esposito2017} and incorporated into the ROS~2 transform tree. Once this calibration is completed, the marker is no longer required. The autonomous inspection process itself is markerless and does not depend on fiducial markers attached to the inspected structure.

\subsection{Pose Logging}

For each acquisition viewpoint, the system stores the captured images together with the corresponding camera pose. The pose is obtained from the ROS~2 transform tree and saved as a translation vector and orientation quaternion. This produces a structured dataset where each image is associated with its spatial pose.

This point is particularly important for neural reconstruction methods. When using methods such as Nerfacto or 3D Gaussian Splatting, the quality of the camera poses strongly influences the quality of the final reconstruction. A coherent and accurately timestamped acquisition pipeline reduces inconsistencies and facilitates the integration of robotic data with reconstruction frameworks.

\section{AUTONOMOUS INSPECTION PIPELINE}

\subsection{Initial LiDAR-Based Volume Estimation}

Before capturing the image dataset, the robot performs an initial scan using the 2D LiDAR. By moving the sensor through a known trajectory and combining the LiDAR readings with the robot pose obtained from the ROS~2 transform tree, a three-dimensional point cloud of the environment is generated.

The resulting point cloud is processed to remove irrelevant points such as floor, ceiling, and distant measurements. Then, a spatial filtering stage is applied to isolate the main target structure. From this filtered point cloud, an approximate bounding box is extracted using Open3D~\cite{zhou2018}. This bounding box is not intended to be a precise reconstruction of the object, but rather a practical geometric approximation used to guide the generation of inspection viewpoints.

\subsection{Viewpoint Generation}

Once the bounding box of the target structure has been estimated, the system generates a set of inspection viewpoints around it. These viewpoints are distributed around the structure in order to maximise visual coverage while keeping the camera within the reachable workspace of the robot.

For each viewpoint, the desired camera orientation is computed so that the optical axis points towards the centroid of the bounding box. This ensures that the target remains centred in the image and increases the usefulness of the captured views for reconstruction.

\begin{figure}[thpb]
        \centering
        \includegraphics[width=1\linewidth]{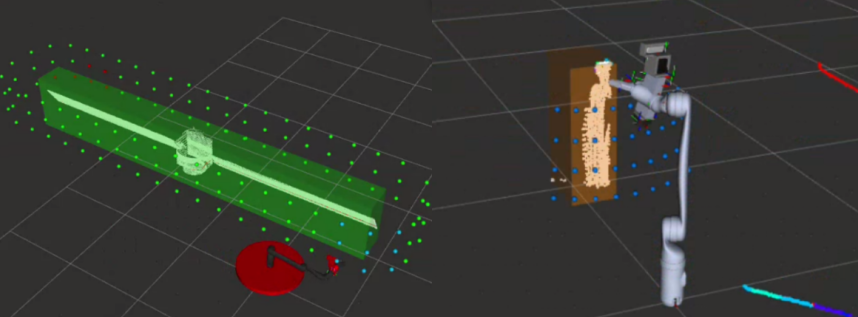}
        \caption{Generated inspection viewpoints around the target structure. The viewpoints are distributed around the estimated inspection volume and oriented towards its centroid, allowing the robot to acquire multiview data for subsequent 3D reconstruction.}
        \label{fig:inspection_viewpoints}
\end{figure}

Fig.~\ref{fig:inspection_viewpoints} illustrates the generated inspection viewpoints around the target structure, showing how the same inspection logic can be represented both in the physical setup and in the virtual environment.

The viewpoint generation process can be adapted depending on the inspection scenario. In simulation, viewpoints can cover the full perimeter of the structure. In the physical laboratory setup, the coverage is limited by the workspace of the fixed manipulator, which restricts the reachable viewpoints to a frontal sector of the structure.

\subsection{Motion Planning}

The motion between inspection viewpoints is planned using MoveIt~\cite{coleman2014}. For each target pose, the inverse kinematics problem is solved and a collision-free trajectory is generated. The planner checks both collisions with the environment and self-collisions of the robot.

In the simulated scenario, the system coordinates the motion of a holonomic base and the robotic arm, increasing the reachable workspace and enabling full coverage around the structure. In the physical setup, the robot operates from a fixed base, making the planning problem more constrained. In this case, MoveIt is especially important for avoiding singularities, respecting joint limits, and finding feasible motions between viewpoints.

\subsection{Synchronized Data Acquisition}

At each inspection viewpoint, the robot stops before capturing data. This reduces motion blur and ensures that the stored pose corresponds to a stable camera configuration. The system then captures RGB, thermal, and depth data, together with the camera pose.

The captured data are stored in a structured directory. Each image is associated with metadata containing the timestamp, translation, orientation, and relevant acquisition parameters. This structure simplifies the conversion of the dataset into the formats required by reconstruction frameworks such as Nerfstudio.

\section{3D RECONSTRUCTION METHOD SELECTION}

Although the objective of this work is autonomous inspection rather than the development of a new 3D reconstruction algorithm, the choice of reconstruction method is important because it affects the quality, usability, and practicality of the inspection output. For this reason, an extensive benchmark of representative reconstruction approaches was conducted, including COLMAP, CasMVSNet, MASt3R, NeuS, 3D Gaussian Splatting through Splatfacto, and Nerfacto. Further background on these methods is provided in Section~II.

The full quantitative benchmark lies outside the main scope of this paper, which focuses on the robotic inspection pipeline. Therefore, the results are simplified in Table~\ref{tab:methods}, where the methods are compared according to criteria relevant for robotic inspection: visual quality, geometric consistency, and ease of integration with the generated robotic dataset.

Based on this analysis, Nerfacto was selected as the reconstruction method used in the experimental validation. This choice was motivated by its balance between reconstruction quality, robustness with posed image datasets, and practical integration within the acquisition pipeline. In this work, the reconstructed model is not considered an end in itself, but an inspection product that can support coverage assessment, visual analysis, and future multimodal diagnosis.

\begin{table}[h]
\caption{Qualitative evaluation of candidate 3D reconstruction methods.}
\label{tab:methods}
\begin{center}
\begin{tabular}{lccc}
\hline
\textbf{Method} & \textbf{Visual quality} & \textbf{Geometry} & \textbf{Integration} \\
\hline
COLMAP & Medium & Medium-High & High \\
CasMVSNet & Medium & Medium & Medium \\
MASt3R & High & Medium & Medium \\
NeuS & Medium & Medium-High & Low \\
3DGS & High & Medium & High \\
Nerfacto & High & Medium-High & High \\
\hline
\end{tabular}
\end{center}
\end{table}

\section{EXPERIMENTAL VALIDATION}

\subsection{Experiment 1: Simulated Inspection with Holonomic Base}

The first experiment was performed in Gazebo using a simulated large space structure. The robotic system consisted of a Kinova Gen~2 manipulator mounted on a holonomic mobile base. This configuration allowed the system to move around the target structure and acquire images from a complete set of viewpoints.

The experiment started with an initial LiDAR scan to estimate the target volume. From the resulting bounding box, the system generated inspection viewpoints around the structure. The motion planner then computed collision-free trajectories between these viewpoints. At each position, the system captured RGB-D data and stored the corresponding camera pose.

\begin{figure}[thpb]
        \centering
        \includegraphics[width=1\linewidth]{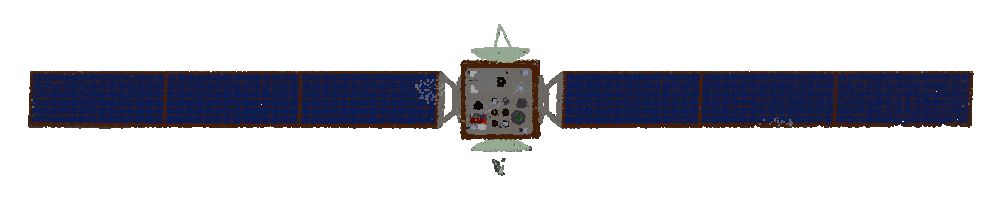}
        \caption{Final reconstruction with the input RGB images from the simulated experiment.}
        \label{fig:simulated_reconstruction}
\end{figure}

The generated dataset was processed using Nerfacto within the Nerfstudio framework~\cite{tancik2023}. As shown in Fig.~\ref{fig:simulated_reconstruction}, the resulting reconstruction confirmed that the selected viewpoints provided sufficient visual coverage and that the stored poses were spatially coherent. The experiment also demonstrated that the simulated pipeline can be used to test inspection strategies before deployment on the physical system. A video demonstration of this experiment is available online.\footnote{\url{https://www.youtube.com/watch?v=ffogW-Ojm5Q}}

\subsection{Experiment 2: Physical Validation with Fixed Base}

The second experiment was carried out in the physical laboratory setup. In this case, the Kinova Gen~2 manipulator was mounted on a fixed base and used to inspect a vertical structure. This scenario introduced more severe workspace limitations than the simulated experiment because the robot could not move around the object.

In consequence, inverse kinematics, singularity avoidance, and motion planning between inspection viewpoints were delegated to MoveIt~\cite{coleman2014}.

Despite these restrictions, the system was able to estimate the approximate structure volume, generate feasible inspection viewpoints, and execute the acquisition sequence. The reachable viewpoints covered approximately the frontal sector of the structure. Although this did not provide full 360-degree coverage, it was sufficient to validate the calibration, data acquisition, and reconstruction pipeline.

\begin{figure}[thpb]
        \centering
        \includegraphics[width=1\linewidth]{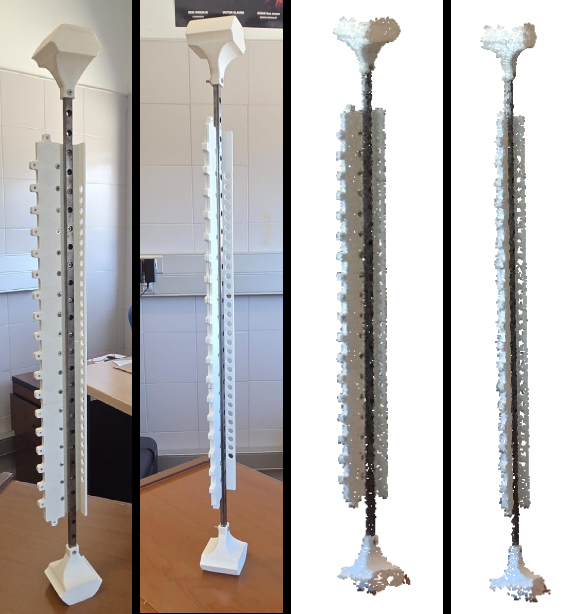}
        \caption{Final reconstruction with the input RGB images from the physical experiment. The two from the left are the real object and the two of the right the reconstruction.}
        \label{fig:physical_reconstruction}
\end{figure}

As shown in Fig.~\ref{fig:physical_reconstruction}, the resulting reconstruction showed that the RGB images and robot poses were coherent enough to generate a 3D model of the inspected object. This confirms that the system can perform autonomous inspection data acquisition under constrained kinematic conditions, producing spatially coherent representations that support subsequent visual and geometric assessment. A video demonstration of the physical experiment is available online.\footnote{\url{https://www.youtube.com/watch?v=XClyWlwP55U}}

\subsection{Thermal-Aware Reconstruction}

To evaluate the contribution of the thermal modality within the proposed inspection platform, an additional reconstruction experiment was performed using Thermal-Nerfacto~\cite{lin2024thermalnerf}, a thermal extension of the Nerfacto pipeline. In this experiment, the synchronized RGB and thermal images acquired by the sensor head were used to train a neural reconstruction capable of rendering both the visual appearance of the scene and its corresponding thermal response.

The objective of this experiment was not to obtain a thermal point cloud, but to evaluate whether the thermal camera can provide an inspection-relevant representation aligned with the reconstructed scene. For inspection purposes, thermal information is mainly useful for identifying regions with abnormal temperature values, such as potential hot spots, overheated components, insulation defects, or thermally degraded areas. Therefore, a visual thermal reconstruction or a set of thermal novel views can be sufficient to support diagnosis, even if the final exported geometry is based primarily on RGB information.

\begin{figure}[thpb]
    \centering
    \includegraphics[width=1\linewidth]{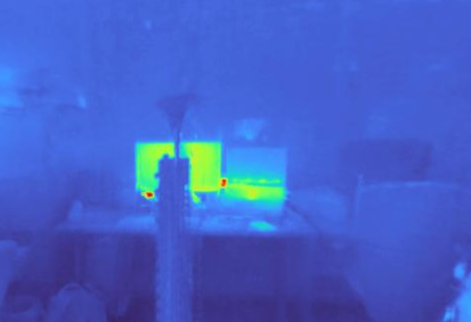}
    \caption{Example of a thermal-aware rendering obtained using Thermal-Nerfacto. The rendered view preserves the thermal contrast of the scene, showing warmer regions that could support future thermal inspection and anomaly analysis.}
    \label{fig:thermal_reconstruction}
\end{figure}

Fig.~\ref{fig:thermal_reconstruction} shows an example of the RGB and thermal outputs obtained with Thermal-Nerfacto. The RGB reconstruction provides the visual and geometric context of the inspected structure, while the thermal rendering highlights the temperature distribution over the observed surface. This result supports the inclusion of the thermal camera in the multimodal sensor head, since it enables future inspection stages based not only on geometry and visual appearance, but also on thermal behaviour.

\subsection{Discussion}

The experiments validate complementary aspects of the proposed architecture. The simulated experiment demonstrates the potential of the system when full coverage around the structure is possible. In this case, the combination of a mobile base and a robotic manipulator allows the system to acquire a complete multiview dataset.

The physical experiment, on the other hand, validates the integration of the real sensors, the calibration procedure, the ROS~2 acquisition pipeline, and the use of real robot poses for reconstruction. Although the fixed-base configuration limits the field of view and reduces the attainable coverage, the experiment confirms that the system can operate under realistic constraints.

The results also highlight the importance of accurate calibration and stable data acquisition. Small inconsistencies in the camera pose, sensor mounting, or intrinsic parameters can affect the final reconstruction. Therefore, the calibration pipeline is not a secondary component, but a central part of the system.

Finally, the Thermal-Nerfacto experiment shows that the synchronized thermal data can be exploited beyond simple image logging. Although the exported geometry is still mainly supported by the RGB reconstruction, the thermal rendering provides an inspection-relevant representation of the temperature distribution of the target. This is particularly useful for future diagnostic tasks, where regions of interest may be defined not only by geometric defects, but also by abnormal thermal behaviour.

The current validation remains preliminary and system-oriented. The simulation demonstrates the complete inspection strategy under increased mobility, whereas the physical experiment validates real sensor integration, calibration, pose logging, motion execution, and reconstruction under the constrained workspace of a fixed-base manipulator. It does not reproduce the complete dynamics, illumination conditions, or operational constraints of an orbital environment.

\section{CONCLUSION}

This paper has presented a markerless autonomous robotic inspection framework for space structures, using 3D reconstruction as an inspection-support representation rather than as the final objective. The system integrates a Kinova Gen~2 manipulator, a multimodal sensor head, ROS~2-based data acquisition, LiDAR-based inspection-volume estimation, automatic viewpoint generation, MoveIt-based motion planning, and synchronized RGB-D, thermal, and pose logging.

The proposed architecture was validated in two complementary robotic scenarios, complemented by an additional thermal-aware reconstruction experiment. First, a Gazebo-based simulation environment with a holonomic base was used to evaluate the inspection strategy under the planar mobility expected from a future ESA microgravity testbed. Second, a physical laboratory experiment with a fixed-base manipulator validated the sensor integration, calibration, pose logging, and data acquisition pipeline under the currently available hardware constraints.

Several 3D reconstruction methods were evaluated as possible inspection-support tools, and Nerfacto was selected due to its balance between visual quality, geometric coherence, and ease of integration with the generated robotic dataset. In addition, Thermal-Nerfacto was used to demonstrate that the synchronized thermal images can be exploited as inspection-relevant renderings aligned with the reconstructed scene. The resulting outputs show that the proposed pipeline can produce spatially coherent inspection data, which is a necessary step towards autonomous inspection of larger and more complex non-cooperative space structures.

Future work will focus on validating the system in a more representative orbital robotics test bench, improving the integration of thermal reconstructions into the inspection process, and developing local high-resolution inspection strategies for regions of potential damage or thermal anomaly. Overall, the proposed framework represents a step towards scalable autonomous robotic inspection of large, non-cooperative, and partially unknown space structures.

\section*{ACKNOWLEDGMENT}

This work was supported by the European Space Agency and Sirin Orbital Systems as part of the ECHO2 project: \textit{End-to-End Control for Handling Operations in Orbit}.


\end{document}